\documentclass[10pt,conference]{IEEEtran}
\usepackage{authblk}
\usepackage[utf8]{inputenc}
\usepackage[hyphens]{url}  
\usepackage{graphicx} 
\usepackage{algorithm}
\usepackage{algorithmic}

\usepackage{newfloat}
\usepackage{listings}

\usepackage{colortbl}
\usepackage[table]{xcolor}
\usepackage{booktabs}   
\usepackage{adjustbox}  
\usepackage{amsmath}
\usepackage{mathtools}
\usepackage{url}

\usepackage{amsfonts}
\usepackage{dsfont}
\usepackage{times}
\usepackage{cleveref}
\usepackage{helvet}
\usepackage{courier}
\usepackage[utf8]{inputenc}
\usepackage[english]{babel}
\usepackage{array} 
\usepackage{lipsum} 

\usepackage{booktabs}

\title{Interpretable Cutting Plane Selection for Mixed-Integer Linear Programming Using Genetic Programming}

\author[1]{Gwen Maudet}
\author[2]{Mathis Da Cunha}
\author[3]{Grégoire Danoy}
\affil[1]{SnT, University of Luxembourg, Esch-sur-Alzette, Luxembourg}
\affil[2]{IRISA, UMR CNRS 6074, F-35700, Rennes, France}
\affil[3]{FSTM/DCS, SnT, University of Luxembourg, Esch-sur-Alzette, Luxembourg}

\begin{document}

\maketitle

\begin{abstract}

Cutting planes (cuts) remove fractional solutions from mixed-integer linear programs (MILPs) without excluding any feasible integer solutions, making them a key ingredient in modern MILP solvers. Current state-of-the-art solvers rely on fine-tuned handcrafted cut strategies, which, despite their effectiveness, remain inherently static and can be suboptimal. More recently, learning-based approaches have been proposed to design adaptive cut strategies; however, most of these methods are based on neural networks, offering limited interpretability and providing no guarantees of convergence.
In this paper, we propose \emph{Genetic Programming and Statistics-based Cut Selection} (GPSCS), the first learning-based framework that explicitly generalizes handcrafted cut selection heuristics into a fully interpretable policy. First, a genetic programming model evolves a symbolic scoring function that ranks candidate cuts using features derived from the cuts and the solving context. Second, a lightweight statistics-based mechanism adaptively determines the number of top-ranked cuts to add at each separation round by detecting significant score dropouts.
We evaluate GPSCS within the SCIP solver on seven classes of synthetic and real-world benchmark instances, with a strong emphasis on generalization: evaluation instances are on average 41 times harder than those used for training. Experimental results show that GPSCS improves average solving time by $6\%$ compared to SCIP's default cut selection strategy and achieves a two-fold speedup over both a recent neural-network-based approach and a no-cut baseline, demonstrating strong robustness and generalization across heterogeneous MILP families.
\end{abstract}


\section{Introduction}

Mixed-Integer Linear Programming (MILP) plays a central role in modeling and solving a wide range of real-world decision-making problems~\cite{archetti_optimization_2022,rais_operations_2011}. Modern MILP solvers like SCIP~\cite{bolusani_scip_2024}, gurobi~\cite{gurobi_gurobi_2024}, CPLEX~\cite{ibm_cplex_ibm_2022} combine multiple complementary components to achieve high solving efficiency. At their core is the branch-and-bound (B\&B) algorithm, which recursively decomposes the original problem into smaller subproblems. Each subproblem is evaluated through its linear programming (LP) relaxation, leading either to a feasible solution of the original MILP if integrality constraints are satisfied, or to a lower bound that can be used to prune the corresponding subtree when a better feasible solution is already known.
A key mechanism for enhancing the effectiveness of the B\&B process is the use of cutting planes (cuts), i.e., valid inequalities that remove fractional LP solutions without excluding any feasible integer solutions. By strengthening the LP relaxation, cuts improve bound quality and, consequently, the solver’s pruning capability. However, the design of an effective cut strategy involves an inherent trade-off: while adding cuts can significantly tighten the LP relaxation, each additional constraint also increases the computational burden of solving the LP, potentially slowing down the overall optimization process. An effective strategy must therefore carefully balance the number of generated cuts and their strengthening power, aiming to select a small subset of highly effective cuts that maximally improve the LP bound.

In practice, cut strategies in modern MILP solvers are largely based on handcrafted heuristics, which use scoring functions defined as linear combinations of cut features to rank candidate cuts and select a fixed number of them depending on the node type (root or non-root)~\cite{achterberg_constraint_2007,suhl_implementing_2012}. While effective in many settings, such heuristics may exhibit highly variable performance depending on the problem structure as the strategy is fixed for all instances.
Recent advances in machine learning (ML) have opened new perspectives for improving decision-making within MILP solvers by enabling the automatic design of adaptive cut strategies~\cite{lodi_learning_2017,bengio_machine_2021,cappart_combinatorial_2021,scavuzzo_machine_2024}. Most existing approaches in this direction rely on neural networks (NN)~\cite{paulus_learning_2022,tang_reinforcement_2020,turner_adaptive_2023,wang_learning_2023}, and have demonstrated significant improvements over static strategies. 
In particular, the work of~\cite{wang_learning_2023} proposes \emph{HEM}, a comprehensive cutting strategy in which, at each node of the B\&B tree, cuts are ranked and a decision is made on the number of cuts to apply. 
Despite their potential effectiveness, these methods suffer from limited interpretability and have no proof of convergence. Related work is discussed in more detail in~\cref{Related_Work}.

In this work, we pursue the same objective of learning a complete cutting strategy that operates at each node of the B\&B tree and is tailored to a specific class of MILP instances. Motivated by recent evidence showing that simple and interpretable learning approaches grounded in existing handcrafted methods can outperform general-purpose NN approaches in combinatorial optimization~\cite{maudet_search_2025}, we adopt a similar perspective.
We introduce \emph{Genetic Programming and Statistics-based  Cut Selection} (GPSCS), an interpretable framework that jointly learns how to rank candidate cuts according to their impact on bound tightening and how many cuts should be added to the formulation. Specifically, GP is used to evolve explicit scoring functions that evaluate cut relevance based on problem-, node-, and cut-level features, thereby generalizing classical handcrafted scoring rules over a much richer and more flexible search space.
Moreover, we propose a lightweight statistics-based dropout detection mechanism that dynamically determines the number of cuts to retain at each node by identifying score breakpoints among the non-parallel cuts. Together, these components define a fully interpretable and adaptive cut strategy, described in detail in~\cref{GP-based_cut}.

We implement our approach within the SCIP framework and evaluate it on seven benchmark classes covering both synthetic and real-world MILP instances. The synthetic benchmarks consist of primal-hard instances, where evaluation problems are significantly larger than those used for training, allowing us to assess generalization within homogeneous problem families. The remaining four classes are drawn from established benchmark libraries and represent heterogeneous real-world instances. Here again, training is performed on simpler instances than those used for evaluation. Overall, evaluation instances require, on average, 41 times more solving time than the corresponding training instances.
We compare GPSCS against several alternatives that combine GP with different strategies for determining the number of retained cuts, including selecting all cuts, only non-parallel cuts, applying statistical dropout detection to all cuts, and a reinforcement-learning (RL)-based cut quantity policy adapted from the HEM method~\cite{wang_learning_2023}. We also evaluate the transferability of the learned GP scoring functions by applying functions trained on one problem class to instances from different classes.
In addition, we compare GPSCS against standard baselines, including the default SCIP cut selection strategy, a configuration without cutting planes (NoCuts), and the neural-network-based cut selection method HEM. 
Among all evaluated methods, GPSCS achieves the best overall performance: it outperforms HEM and \emph{NoCuts} with an average two-fold speedup and improves upon SCIP's default cut selection strategy by $6\%$ on average. Detailed experimental results are presented in~\cref{Experiments}.

The main contributions of this paper are summarized as follows:

\par\noindent$\bullet$ We define a general, feature-based cut scoring space that encompasses and extends traditional handcrafted scoring rules.
\par\noindent$\bullet$ We propose a GP-based method to automatically  explore this space to generate interpretable cut scoring functions trained to be efficient on training instances.
\par\noindent$\bullet$  We introduce a simple statistics-based procedure to adaptively determine the number of cuts to apply by detecting breakpoints in the score distribution.
\par\noindent$\bullet$  We conduct an extensive empirical evaluation on seven benchmark classes, including both synthetic and real-world instances.
\par\noindent$\bullet$  We experimentally demonstrate the effectiveness of our adaptive cut number selection by comparing it against four alternative strategies for choosing the number of cuts.
\par\noindent$\bullet$  GPSCS achieves significant performance improvements over the default solver strategy, a NoCut configuration, and the state-of-the-art NN-based method HEM, while also highlighting potential convergence issues in HEM on certain instance classes.


\section{Background}\label{Background}

This section presents the key concepts underlying our approach. We first review the B\&B algorithm, which forms the foundation of modern MILP solvers. Next, we discuss cutting planes and the strategies for selecting them. Finally, we introduce GP, a method that evolves a population of programs over successive generations.

\subsubsection{B\&B}

A MILP $P$ is defined by decision variables $x = (x_i)_{i \in \{1,\dots,n\}}$, where a subset indexed by $\mathcal{J} \subseteq \{1,\dots,n\}$ is required to take integer values. The objective is to minimize a linear function $c^\top x$ subject to $m$ linear constraints.
MILPs are typically solved using the B\&B algorithm, which recursively partitions the feasible region into subproblems organized as a search tree. At each node $N$, integrality constraints are relaxed, yielding an LP relaxation that can be solved efficiently. Let $x^{LP}$ denote an optimal LP solution with objective value $z^{LP} = c^\top x^{LP}$.
If $x^{LP}$ satisfies all integrality constraints, it is optimal for the subproblem and updates the best known feasible objective value $z^*$.  Otherwise, $z^{LP}$ provides a valid lower bound on the subproblem. If this bound is no better than the incumbent solution (i.e., $z^{LP} \geq z^*$), the entire subtree rooted at $N$ can be \emph{pruned}. If not, the subproblem is branched into two or more child subproblems whose union covers the feasible region of the parent. This process continues until all nodes are either solved or pruned. The efficiency of B\&B critically depends on its ability to prune large portions of the search tree early.

\subsubsection{Cut Selection Strategy}
As the B\&B’s efficiency largely derives from its pruning capabilities, it depends on the quality of the bounds provided by LP relaxations. Cutting planes are introduced to strengthen these bounds. A \emph{cut} is a valid linear inequality of the form $r:\alpha^\top x \leq \beta$ that eliminates fractional LP solutions without excluding any feasible integer solution. By tightening the LP relaxation, cuts increase the likelihood of early pruning during the search.
Cuts are incorporated through an iterative procedure where for each B\&B node, candidate cuts are generated by dedicated cut generators (such as Gomory cut generation~\cite{gilmore_linear_1961} and others), a subset is selected, and the LP is reoptimized with the additional constraints. This process defines the \emph{cut strategy}, which can be characterized by two main components~\cite{dey_theoretical_2018}: (i) a ranking or scoring mechanism used to assess the relevance of candidate cuts, and (ii) a rule determining the number of cuts ultimately added to the formulation. While adding cuts generally strengthens the LP relaxation, including too many cuts can substantially increase the size of the formulation and degrade LP solving performance. Further background on MILP solving, B\&B, and cutting planes can be found in~\cite{achterberg_constraint_2007}.

\subsubsection{GP}
Evolutionary algorithms are population-based metaheuristics inspired by the principles of natural evolution. They operate by maintaining a population of candidate solutions, referred to as \emph{individuals}, which is iteratively evolved over successive \emph{generations} according to a fitness criterion that quantifies solution quality with respect to the objective of interest.
A distinctive characteristic of GP is that each individual encodes a computer program, most commonly represented as a tree structure. Leaf nodes, called \emph{terminals}, correspond to the basic elements of the program, while internal nodes represent \emph{operators} that combine these elements. The evolutionary process of an evolutionary algorithm starts from an initialized population and, at each generation, produces a new population through the application of three main operators: (i) \emph{selection}, which favors individuals with higher fitness for reproduction; (ii) \emph{crossover}, which combines two or more individuals by exchanging parts; and (iii) \emph{mutation}, which introduces random modifications to individuals. This process continues until a predefined stopping condition is met, such as a maximum number of generations. The population-based and stochastic nature of GP allows all individuals within a generation to be evaluated independently, enabling embarrassingly parallel execution and making GP particularly well suited for large and unstructured search spaces. 
When this space corresponds to a space of heuristics, GP can be viewed as a generative hyper-heuristic~\cite{dokeroglu_hyper-heuristics_2024}. A comprehensive overview of evolutionary algorithms is provided in~\cite{talbi_metaheuristics_2009}.

 

\section{Related Work}\label{Related_Work}

This section reviews prior work on cut selection strategies. We first discuss handcrafted heuristics that are still widely used as default strategies in modern solvers. We then survey recent ML-based approaches that aim to learn adaptive cut selection policies. Finally, we position our work as a lightweight and interpretable alternative that leverages GP to generalize handcrafted scoring rules and combines it with a simple drop detection for determining the number of cuts.

\subsubsection{Handcrafted Heuristics}

Modern MILP solvers such as SCIP, Gurobi, and CPLEX~\cite{bolusani_scip_2024,gurobi_gurobi_2024,ibm_cplex_ibm_2022} typically rely on handcrafted scoring functions to evaluate and rank candidate cuts. For instance, SCIP defines its scoring function as a weighted combination of four primary cut quality metrics: (i) \emph{efficacy} (eff), measuring the distance between the cut and the LP solution, defined as $(\alpha^\top x^{LP} - \beta) / \lVert \alpha \rVert$; (ii) \emph{directed cutoff distance} (dcd), which incorporates information from the best known feasible solution, defined as $(\alpha^\top x^{LP} - \beta) / \lvert \alpha^\top y \rvert$, with $y = (x^* - x^{LP}) / \lVert x^* - x^{LP} \rVert$; (iii) \emph{objective parallelism} (obp), capturing the alignment between the cut and the objective function vector, defined as $\lvert \alpha^\top c / (\lVert \alpha \rVert \lVert c \rVert) \rvert$; and (iv) \emph{integer support} (isp), measuring the involvement of integer-constrained variables, defined as $N_{\text{int}} / N_{\text{nonz}}$, where $N_{\text{nonz}} = \sum_{i=1}^n \mathds{1}(\alpha_i \neq 0)$ and $N_{\text{int}} = \sum_{i \in \mathcal{J}} \mathds{1}(\alpha_i \neq 0)$, with $\mathds{1}$ denoting the indicator function. Another important metric is the \emph{normalized violation} (nvi)~\cite{wang_learning_2023,huang_learning_2021}, defined as $(\alpha^\top x^{LP} - \beta)/|\beta|$.  
Empirical and theoretical studies indicate that using fixed global weights for these metrics is suboptimal on heterogeneous instance distributions~\cite{turner_adaptive_2023}, motivating the need for adaptive, instance-aware cut scoring strategies.

Similarly, the number of cuts added per separation round is typically governed by fixed limits; in SCIP, for instance, larger limits are applied at the root node and smaller limits elsewhere, often retaining most cuts. Procedures are also used to discard nearly parallel cuts to preserve effectiveness. However, as highlighted in~\cite{dey_theoretical_2018,wang_learning_2023}, determining the appropriate number of cuts is critical, and fixed limits are generally suboptimal, motivating adaptive strategies that balance LP strengthening with computational overhead.

\subsubsection{ML Approaches}
Recent advances in ML have led to the formulation of several solver decisions as learning problems, including node selection, branching, and cut selection, as summarized in several surveys~\cite{lodi_learning_2017,bengio_machine_2021,cappart_combinatorial_2021,scavuzzo_machine_2024}. These approaches can be broadly categorized into imitation learning and self-learning.
For cut selection,~\cite{paulus_learning_2022} propose an imitation learning approach in which a NN is trained to mimic an oracle that selects the best scoring metric from a predefined set of handcrafted measures, thereby reducing the computational cost of oracle-based decisions at inference time. However, this method only selects a single cut per separation round, whereas modern solvers typically add multiple cuts simultaneously to achieve stronger LP tightening. Moreover, such an oracle can only be heuristic in nature, which inherently bounds the performance of the learned model by the quality of the oracle itself and may therefore lead to suboptimal strategies. For these reasons, most recent work has focused on self-learning approaches based on RL. One of the earliest applications in this direction is proposed in~\cite{tang_reinforcement_2020}, where deep RL is used to select Gomory cuts, again restricting the selection to a single cut per round, which limits practical applicability. The approach proposed in~\cite{turner_adaptive_2023} extends this line of work by enabling the selection of multiple cuts through the dynamic adjustment of the weights used in the cut scoring mechanism of SCIP. However, the number of cuts added remains fixed, and the learned policy is evaluated solely based on bound improvements at the root node.
A more comprehensive proposition of cut selection is given with \emph{HEM} in~\cite{wang_learning_2023}, which jointly addresses cut ranking and cut quantity decisions that are done along all the solving process. The authors introduce a hierarchical  NN based RL framework composed of a high-level policy that determines the number of cuts to add and a lower-level sequence model that ranks candidate cuts. Although this approach improves upon static solver configurations, the resulting policies are black-box models, limiting interpretability.
More importantly, the absence of prior structural or heuristic knowledge requires exploring a greatly enlarged decision space, combining both the number of cuts and a total ordering over all candidates. Because the reward is the solving time after fully exploring the B\&B tree, this large space coupled with sparse, delayed feedback makes training highly demanding, potentially unstable, and without guaranteed convergence.

\subsubsection{Positioning}

While NN-based approaches—particularly those relying on self-learning—offer a general and powerful framework for modeling solver decisions, their all-purpose nature often entails expensive training times and provides no guarantee of stable or rapid convergence toward high-quality policies. In contrast, leveraging decades of domain knowledge embedded in handcrafted heuristics to guide automated method generation enables faster convergence, stronger inductive bias, and the production of interpretable strategies.

GP provides a lightweight and interpretable alternative to NN-based methods by directly evolving executable and human-readable heuristics. Recent work~\cite{maudet_search_2025} demonstrates that GP can be used as a hyper-heuristic to learn effective B\&B search strategies by generalizing existing handcrafted heuristics through symbolic expressions, achieving improvements over default solver configurations and NN methods, while requiring significantly less computational effort. Our work builds on this idea by applying GP to the cut selection problem, which has not previously been explored using GP. In contrast to NN-based RL approaches, our framework produces lightweight scoring functions that are evaluated at each B\&B node. These functions combine features derived from the cut, the current node, and the problem instance, and extend classical handcrafted scoring rules over a substantially richer search space, enabling the discovery of novel and effective cut ranking policies that generalize across instance classes.

Beyond cut ranking, the number of cuts added at each separation round and node has a significant impact on solver performance~\cite{dey_theoretical_2018,wang_learning_2023}. Rather than learning this decision through an additional learning procedure that would imply building a looping procedure to ensure convergences with the GP, we propose a simple deterministic cutoff method based on statistical analysis of score distributions. By assuming that the scoring function reflects cut quality, sharp drops in the distribution signal lower-ranked cuts unlikely to add value; our method then retains only the highest-scoring cuts. In our framework, the GP component ensures this accurate scoring, thereby guaranteeing the effectiveness of the statistics-based cutoff.

\section{GP-based Cut Selection Strategy}\label{GP-based_cut}

We introduce \emph{genetic programming and statistics-based cut selection} (GPSCS), a learning-based cut strategy built around two complementary components. First, we propose a selection procedure that filters highly parallel cuts and adaptively determines how many cuts should be added at each separation round. Second, we leverage GP to automatically learn a cut scoring function that ranks candidate cuts based on their expected impact.
In contrast to traditional solver heuristics, which rely on manually designed and fixed scoring rules, GPSCS learns an explicit and fully interpretable strategy that adapts both across problem instances and dynamically across nodes of the B\&B tree during the solving process.

\subsection{Selection Procedure According to a Cut Scoring Function}

We first describe the procedure used to determine the final subset of cuts to be applied at each separation round and at each node, given a cut scoring function.
Let $f(r,N,P)$ denote a scoring function that assigns a numerical value to a cut $r$ generated at node $N$ of the B\&B tree for a problem instance $P$. Evaluating $f$ over the set of generated cuts induces an ordering $\{(r_i)_{1 \leq i \leq l}\}$ such that $i \leq j$ implies $f(r_i) \geq f(r_j)$. The objective is to select a subset $\mathcal{R} \subseteq \{r_1,\dots,r_l\}$ of cuts to be added to the formulation.
The selection procedure consists of two successive stages. First, a filtering step removes highly parallel cuts to promote geometric diversity. Second, a dropout detection heuristic is applied to the filtered list in order to identify a natural cutoff point in the ranking, retaining only the most promising cuts.

\subsubsection{Prioritization of Orthogonal Cuts}

Starting from the initial ranking induced by the learned scoring function, we apply a filtering procedure commonly used both in solvers and in recent research~\cite{bolusani_scip_2024,turner_adaptive_2023}. The objective is to prioritize cuts that introduce genuinely new geometric information, rather than reinforcing directions already well represented by previously selected cuts. Such a criterion has been consistently shown to improve the effectiveness of cutting planes by reducing redundancy and strengthening LP relaxations more efficiently~\cite{achterberg_constraint_2007,dey_theoretical_2018,turner_adaptive_2023,wang_learning_2023}.
Since cut parallelism can be computed explicitly, handling this filtering step separately simplifies the learning task faced by the scoring function. 

Concretely, cuts are processed iteratively from highest to lowest score. For each selected cut $r:\alpha^\top x \leq \beta$, the absolute cosine similarity with every lower-ranked cut $r':{\alpha'}^\top x \leq \beta'$ is computed as $s(r,r') = |\alpha^\top \alpha'| / (\lVert \alpha \rVert \lVert \alpha' \rVert)$, where $s(r,r')=1$ denotes perfect parallelism and $s(r,r')=0$ orthogonality. Any lower-ranked cut whose similarity exceeds a predefined threshold (set to $0.9$ in our experiments) is discarded.



\subsubsection{Heuristic for Determining the Number of Cuts}

After filtering highly parallel cuts from the ranked list, the remaining question is how many of the remaining candidates should be retained at the current separation round. We address this question through a score dropout detection heuristic, based on the assumption that the learned scoring function induces a meaningful ordering of cut quality: a pronounced drop between consecutive scores indicates diminishing marginal benefits from adding further cuts. Since adding cuts is generally beneficial, the goal is not aggressive filtering but rather removing only the lowest-quality cuts at the bottom of the ranking.

Let $\{(r_i)_{1 \leq i \leq l}\}$ denote the filtered and ordered list of cuts, and let $\Delta = \{ |f(r_i) - f(r_{i+1})| \}_{1 \leq i \leq l-1}$ be the sequence of absolute differences between consecutive scores. Since scoring functions may operate on different numerical scales, these differences are normalized using a z-score~\cite{norman_pdq_2003}, by computing for each $\Delta_i$ the value $z_i = (\Delta_i - \operatorname{mean}(\Delta)) / \operatorname{std}(\Delta)$. 
We aim to detect a significant gap near the end of the ranked list. To this end, we introduce a threshold $z_{\max}$ and scan the sequence $\{z_i\}$ for large gaps. If one or more indices satisfy $z_i > z_{\max}$, we select the largest such index; otherwise, we select the index of the largest observed gap. Formally, the stopping index is $i_{\text{threshold}} = \max\!\left( \{ i \mid z_i > z_{\max} \} \right)$ if non-empty, and $i_{\text{threshold}} = \arg\max_i z_i$ otherwise. Only the first $i_{\text{threshold}}$ cuts in the ranked list are added to the formulation.

\subsection{GP-based Cut Scoring Method}

We now present the GP framework for learning cut scoring functions, first defining the policy space and then describing how it is explored.

\subsubsection{Definition of the Cut Scoring Policy Space}

We define the cut scoring policy space as the set of symbolic expressions generated by combining cut- and context-related variables with arithmetic operators. Each expression assigns a score to every candidate cut, thereby inducing a ranking from the most to the least promising. Our objective is to generalize the scoring strategies used in modern MILP solvers by defining a richer policy space that subsumes existing handcrafted heuristics while exploiting additional information about the current B\&B node and the problem instance.

As terminal symbols, we include the classical cut quality features $eff$, $dcd$, $obp$, $N_{\text{int}}$, $N_{\text{nonz}}$, and $nvi$. To capture the local solving context, we additionally include the node depth $d_i$, while the number of variables $n$ and constraints $m$ encode instance-level characteristics. Finally, a large positive constant $M$ enables strong prioritization when required.
The operator set consists of the arithmetic operators $+, -, \times, \div$ together with the unary operators $\exp(\cdot)$ and $\sqrt{\cdot}$, allowing both linear and non-linear scoring functions. For instance, a policy prioritizing efficacy while favoring sparse cuts is given by $f(r) = eff(r) \times M - N_{\text{nonz}}(r)$.

\subsubsection{GP Framework}
We employ GP to explore the space of cut scoring heuristics introduced above, thereby framing our approach as a generative hyper-heuristic~\cite{dokeroglu_hyper-heuristics_2024}. We adopt a standard GP framework~\cite{talbi_metaheuristics_2009}, in which a population of $p$ individuals—each encoding a candidate scoring function—is evolved over a fixed number of generations $N$. The initial population is generated randomly, with tree depths bounded between $D_{\text{init-min}}$ and $D_{\text{init-max}}$.

The population evolves across generations according to a fitness-driven selection mechanism. In our setting, the objective is to learn a scoring function that efficiently solves a set of training set of instances belonging to the same problem class. Since all training instances are solved to optimality, the most reliable and least biased fitness measure is the solving time. We therefore define the fitness of an individual as the average solving time over the training set, with a time limit of $500$ seconds assigned as a penalty when the limit is reached.

For selection, we employ a double tournament strategy~\cite{luke_fighting_2002} designed to limit excessive tree growth and mitigate overfitting. In the first tournament, $k$ individuals are sampled uniformly at random and the one with the best fitness is selected. In the second tournament, two winners from the first stage are compared based on tree size, and the shallower tree is selected with probability $p_{\text{pars}}/2$, where $p_{\text{pars}} \in [1,2]$.

After selection, crossover is applied with probability $P_{\text{mate}}$ (typically above $0.8$). We use one-point subtree crossover: a node is randomly selected in each parent tree, and the corresponding subtrees are exchanged to produce two offspring. Mutation is then applied independently to each individual with probability $P_{\text{mut}}$ (typically below $0.15$). During mutation, a randomly selected subtree is replaced by a newly generated subtree whose depth lies between $D_{\text{mut-min}}$ and $D_{\text{mut-max}}$.

\subsubsection{Training and Deployment}

During training, each GP-derived scoring function is evaluated by solving all training instances to optimality or until the time limit is reached. For each instance and at each B\&B node, candidate cuts are ranked using the GP-based scoring function, filtered to promote orthogonality, and selected using the dropout detection heuristic described earlier. At deployment time, the learned scoring function is applied unchanged, following the exact same ranking, filtering, and selection procedure, ensuring full consistency between training and inference.



\section{Experiments}\label{Experiments}
This section presents an experimental evaluation of our approach. We first assess several variants of GPSCS by considering different strategies for determining the number of retained cuts and by evaluating GP scoring functions trained on classes different from those used for testing. We then compare GPSCS against established cut selection strategies, including a recent learning-based method. Experiments are conducted on seven classes of instances. The first three classes are synthetically generated, with evaluation instances significantly larger than those used for training, in order to assess the generalization capability of the proposed method. The remaining four classes are drawn from state-of-the-art benchmarks and consist of heterogeneous, real-world instances, with training performed on simpler instances than those in evaluation, allowing us to test performance on realistic problem distributions.

\subsubsection{Experimental Setup}

All methods are implemented within the open-source solver SCIP version~9.2.2, with all default solver components left unchanged except for the cut selection strategy. The source code is publicly available at \url{https://......git}. For both training and evaluation, each run is executed on a dedicated node of an HPC cluster. CPU-only runs are carried out on nodes equipped with 128 cores and 256\,GB of RAM, while runs involving GPU resources use nodes with 28 CPU cores, 768\,GB of RAM, and four NVIDIA V100 GPUs with 16\,GB each. Since SCIP itself solves each instance using a single CPU core, allocating an entire node primarily provides access to large memory capacities; in particular, GPU-equipped nodes offer a hardware advantage in terms of available RAM and additional GPUs. These multi-core and multi-GPU resources are exploited through parallelism during the training phase, whereas evaluation instances are solved sequentially, one at a time. To prevent excessive computation times, we focus on instances that can be solved within a reasonable duration and impose a time limit of 500 seconds per instance.

Experiments are conducted on seven instance classes, each split into $50$ instances for training and $50$ for evaluation. The same training sets are used for all methods requiring training (all GP variants and the NN-based HEM).  
The first three classes are synthetically generated to evaluate generalization to larger instance from homogeneous training set: (i) Generalized Independent Set (GISP)~\cite{chmiela_learning_2021}, with training graphs of size $n_g \in [60,70]$ and evaluation graphs of size $n_g \in [120,140]$; (ii) Maximum Satisfiability (MAXSAT)~\cite{bejar_generating_2009}, using the same size scaling; and (iii) Fixed Charge Multicommodity Network Flow (FCMCNF)~\cite{hewitt_combining_2009}, with $n_g = 15$ nodes for training and $n_g = 20$ for evaluation, and $m = 1.5 n_g$ commodities~\cite{chmiela_learning_2021}. Graphs are generated via the Erdos-Renyi model~\cite{erdos_evolution_1961} with edge probabilities $0.3$ for GISP and MAXSAT, and $0.6$ for FCMCNF.  
The remaining four classes are drawn from established benchmarks to evaluate performance on heterogeneous, real-world instances, also using easier instances for training and harder ones for evaluation. For selection, we first choose 100 instances per class that can be solved within 300 seconds, then use the 50 fastest for training and the remaining 50 for evaluation. These include the Mixed-Integer Knapsack Polyhedron (MIK) class~\cite{Atamturk_facets_2003} and the three StrIPlib~\cite{bastubbe_striplib_2025} classes with the largest number of instances—binpacking, cuttingstock, and capacitated lot-sizing—to ensure sufficient heterogeneity.

We compare our approach against three baselines: a configuration without cutting planes (\emph{NoCuts}), the default SCIP cut selection strategy (\emph{SCIP}), and a recent neural-network-based method that jointly learns cut ranking and cut quantity decisions~\cite{wang_learning_2023} (\emph{HEM}). 
For the GP-based methods, we investigate several alternatives 
by considering the following configurations: (i) \emph{GPSCS}, our proposed method applied to the list of non-parallel cuts only, where the statistics-based decision rule determines the number of retained cuts for non parallel cuts; (ii) \emph{GPSAll}, which applies the same decision rule to the complete set of generated cuts. To favor geometric diversity, non-parallel cuts are ranked first, followed by the filtered parallel cuts, each group remaining internally ordered according to the GP score. The score-dropout detection is applied independently within each group, preventing the transition between the two groups from influencing the cutoff decision; (iii) \emph{GPAll}, which retains all generated cuts; (iv) \emph{GPNP}, which retains all non-parallel cuts; and (v) \emph{GPRL}, a hybrid variant that uses the GP scoring function learned by \emph{GPSCS} while replacing the statistical decision rule with the RL-based cut quantity module of \emph{HEM}. All these variants follow the same pipeline: the specific cut ranking function is incorporated during GP training, producing a distinct GP-derived scoring function for each alternative. Finally, to evaluate the transferability of GP scoring functions across problem classes, we consider \emph{WGPSCS}. For each evaluation instance, \emph{WGPSCS} randomly selects a GP scoring function trained on a different problem class and combines it with the same statistics-based decision rule as \emph{GPSCS}.
For \emph{GPSCS}, \emph{WGPSCS} and \emph{GPSAll}, the dropout-based cut quantity mechanism uses a maximum z-score threshold of $z_{\max}=1$. All GP trainings employ a population of $p=64$ individuals evolved over $N=20$ generations, with fitness evaluations performed in parallel on a CPU node. Selection is based on tournament selection with size $k=5$ and a parsimony coefficient of $p_{\text{pars}}=1.2$, while crossover and mutation are applied with probabilities $P_{\text{mate}}=0.9$ and $P_{\text{mutate}}=0.1$, respectively. Initial tree depths are bounded between 1 and 17, and mutation-generated subtrees are restricted to depths between 1 and 5.
Since GP is inherently stochastic, for all GP-based variants, we perform 15 independent training runs with different seeds and retain the scoring function achieving the best performance on the training set. For the NN baseline, the \emph{HEM} approach needs training two neural networks in parallel over 100 epochs on four GPUs of a single computing node.

\subsubsection{Performance Evaluation}



\begin{table}[htbp]
\centering
\footnotesize
\renewcommand{\arraystretch}{1.5}
\begin{tabular}{l|m{3.5cm}|c}
\textbf{Problem} & \textbf{GP Function} & \textbf{Train perf} \\
\hline
FCMCNF & $-\left(m + \dfrac{\exp(\exp(-N_{\text{int}}))}{eff}\right)$ & $3.53\pm 0.10$ \\
MAXSAT & $nvi \times (dcd - d_i)$ & $4.25\pm 0.06$ \\
GISP & $\dfrac{d_i - \sqrt{eff}}{nvi}$ & $2.53\pm 0.02$ \\
MIK & $\exp(dcd \times m)$ & $16.65\pm 0.48$ \\
Cuttingstock & $\left(dcd - \left(\dfrac{m}{dcd} + obp\right)\right) \times n$ & $7.12\pm 1.40$ \\
Binpacking & $-\dfrac{nvi}{dcd \times eff}$ & $6.83\pm 0.14$ \\
Cap lot-sizing & $-\dfrac{N_{\text{nonz}}}{dcd \times eff}$ & $4.19\pm 0.02$ \\
\end{tabular}
\caption{Best GPSCS GP functions selected from 15 runs based on training solving time; values report the mean and standard deviation of the selected functions across runs.}
\label{table_scoring_function}
\end{table}










\renewcommand{\arraystretch}{1}
\begin{table*}[htbp]
\centering
\footnotesize
\begin{tabular}{@{}lccccccc|c@{}}
\hline
Method & FCMCNF & MAXSAT & GISP & MIK & Binpacking & Cuttingstock & Cap lot-sizing & Mean\\
\hline

NoCuts& \textcolor[cmyk]{0.00,1.00,1,0.25}{$500.0\pm0$}& \textcolor[cmyk]{0.93,0.07,1,0.02}{$262.2\pm202$}& \textcolor[cmyk]{1.00,0.00,1,0.00}{$71.3\pm85$}& \textcolor[cmyk]{0.00,1.00,1,0.25}{$234.7\pm215$}& \textcolor[cmyk]{0.27,0.73,1,0.18}{$293.2\pm189$}& \textcolor[cmyk]{0.00,1.00,1,0.25}{$452.3\pm124$}& \textcolor[cmyk]{0.00,1.00,1,0.25}{$500.0\pm0$}& \textcolor[cmyk]{0.00,1.00,1,0.25}{$330.5\pm116$}\\
SCIP& \textcolor[cmyk]{0.89,0.11,1,0.03}{$25.7\pm23$}& \textcolor[cmyk]{0.85,0.15,1,0.04}{$281.6\pm170$}& \textcolor[cmyk]{0.00,1.00,1,0.25}{$334.2\pm107$}& \textcolor[cmyk]{1.00,0.00,1,0.00}{$14.5\pm10$}& \textcolor[cmyk]{1.00,0.00,1,0.00}{$169.1\pm54$}& \textcolor[cmyk]{0.59,0.41,1,0.10}{$181.0\pm43$}& \textcolor[cmyk]{1.00,0.00,1,0.00}{$147.9\pm53$}& \textcolor[cmyk]{0.95,0.05,1,0.01}{$164.9\pm66$}\\
HEM& \textcolor[cmyk]{0.00,1.00,1,0.25}{$500.0\pm0$}& \textcolor[cmyk]{0.18,0.82,1,0.21}{$447.1\pm111$}& \textcolor[cmyk]{0.00,1.00,1,0.25}{$204.3\pm77$}& \textcolor[cmyk]{0.00,1.00,1,0.25}{$295.0\pm221$}& \textcolor[cmyk]{0.00,1.00,1,0.25}{$400.2\pm175$}& \textcolor[cmyk]{0.00,1.00,1,0.25}{$399.9\pm192$}& \textcolor[cmyk]{0.00,1.00,1,0.25}{$500.0\pm0$}& \textcolor[cmyk]{0.00,1.00,1,0.25}{$392.4\pm111$}\\
\hline
GPRL& \textcolor[cmyk]{0.83,0.17,1,0.04}{$27.1\pm22$}& \textcolor[cmyk]{0.85,0.15,1,0.04}{$283.2\pm174$}& \textcolor[cmyk]{0.00,1.00,1,0.25}{$399.7\pm145$}& \textcolor[cmyk]{0.00,1.00,1,0.25}{$33.4\pm34$}& \textcolor[cmyk]{0.67,0.33,1,0.08}{$225.6\pm160$}& \textcolor[cmyk]{0.62,0.38,1,0.09}{$176.4\pm184$}& \textcolor[cmyk]{0.28,0.72,1,0.18}{$255.1\pm138$}& \textcolor[cmyk]{0.72,0.28,1,0.07}{$200.1\pm122$}\\
GPAll& \textcolor[cmyk]{0.75,0.25,1,0.06}{$28.8\pm23$}& \textcolor[cmyk]{0.74,0.26,1,0.07}{$309.4\pm170$}& \textcolor[cmyk]{0.00,1.00,1,0.25}{$292.6\pm107$}& \textcolor[cmyk]{0.17,0.83,1,0.21}{$26.5\pm22$}& \textcolor[cmyk]{0.82,0.18,1,0.05}{$199.8\pm170$}& \textcolor[cmyk]{0.57,0.43,1,0.11}{$182.8\pm158$}& \textcolor[cmyk]{0.64,0.36,1,0.09}{$201.1\pm105$}& \textcolor[cmyk]{0.87,0.13,1,0.03}{$177.3\pm108$}\\
GPNP& \textcolor[cmyk]{0.96,0.04,1,0.01}{$24.1\pm18$}& \textcolor[cmyk]{0.86,0.14,1,0.04}{$280.2\pm168$}& \textcolor[cmyk]{0.00,1.00,1,0.25}{$321.6\pm123$}& \textcolor[cmyk]{0.60,0.40,1,0.10}{$20.3\pm19$}& \textcolor[cmyk]{0.88,0.12,1,0.03}{$188.8\pm160$}& \textcolor[cmyk]{1.00,0.00,1,0.00}{$128.1\pm159$}& \textcolor[cmyk]{0.79,0.21,1,0.05}{$178.7\pm117$}& \textcolor[cmyk]{0.96,0.04,1,0.01}{$163.1\pm109$}\\
GPSAll& \textcolor[cmyk]{0.97,0.03,1,0.01}{$23.8\pm17$}& \textcolor[cmyk]{0.87,0.13,1,0.03}{$276.3\pm165$}& \textcolor[cmyk]{0.00,1.00,1,0.25}{$301.8\pm119$}&\textcolor[cmyk]{0.68,0.32,1,0.08}{$19.1\pm15$}& \textcolor[cmyk]{0.75,0.25,1,0.06}{$211.9\pm155$}& \textcolor[cmyk]{0.87,0.13,1,0.03}{$144.7\pm165$}& \textcolor[cmyk]{0.71,0.29,1,0.07}{$190.8\pm107$}& \textcolor[cmyk]{0.93,0.07,1,0.02}{$166.9\pm106$}\\
GPSCS& \textcolor[cmyk]{1.00,0.00,1,0.00}{$23.1\pm17$}& \textcolor[cmyk]{1.00,0.00,1,0.00}{$245.2\pm150$}& \textcolor[cmyk]{0.00,1.00,1,0.25}{$306.3\pm100$}& \textcolor[cmyk]{0.71,0.29,1,0.07}{$18.7\pm14$}& \textcolor[cmyk]{0.87,0.13,1,0.03}{$191.4\pm159$}& \textcolor[cmyk]{0.93,0.07,1,0.02}{$137.4\pm139$}& \textcolor[cmyk]{0.83,0.17,1,0.04}{$172.4\pm117$}& \textcolor[cmyk]{1.00,0.00,1,0.00}{$156.4\pm99$}\\
\hline
WGPSNP& \textcolor[cmyk]{0.00,1.00,1,0.25}{$500.0\pm0$}& \textcolor[cmyk]{0.27,0.73,1,0.18}{$423.4\pm124$}& \textcolor[cmyk]{0.00,1.00,1,0.25}{$269.1\pm103$}&\textcolor[cmyk]{0.00,1.00,1,0.25}{$258.7\pm218$}& \textcolor[cmyk]{0.25,0.75,1,0.19}{$296.6\pm190$}& \textcolor[cmyk]{0.00,1.00,1,0.25}{$450.3\pm130$}& \textcolor[cmyk]{0.00,1.00,1,0.25}{$500.0\pm0$}& \textcolor[cmyk]{0.00,1.00,1,0.25}{$385.4\pm109$}\\

\hline
\end{tabular}

\caption{Comparative results across seven benchmark classes for three state-of-the-art cut selection strategies and five GP-based variants. Reported values are the mean (standard deviation) of solving times. Green highlights the best baseline per class, red indicates solving times at least two times longer than the best, with a gradient used for intermediate values.}
\label{perf_table}
\end{table*}




The retained scoring functions learned by \emph{GPSCS}, namely the candidates achieving the lowest average solving time on the training set, are reported in~\cref{table_scoring_function}. We also summarize the overall GP training behavior by reporting, across the 15 independent runs, the mean solving time and standard deviation of the selected best-performing candidates. Although not shown explicitly, the GP runs do not converge to identical symbolic expressions. Nevertheless, their performance is highly consistent: for the synthetically generated instances, the standard deviation remains below $0.06$ seconds for mean solving times of only a few seconds, while slightly larger variations are observed on benchmark classes due to their higher instance heterogeneity.
Convergence plots leading to the final \emph{GPSCS} scoring functions are provided in Appendix~C. While the learned functions differ substantially across problem classes, highlighting the need for problem-specific cut selection strategies, some cut quality features consistently appear across the learned policies. Appendix~D provides a frequency-based analysis of feature occurrences and their contribution to score variations, distinguishing whether increasing a feature value tends to increase or decrease the resulting cut score.

\cref{perf_table} reports the mean solving times of the three baselines and the five GP-based variants across the seven problem classes. Evaluation instances are on average 41 times harder than training instances in terms of solving time, highlighting the significant generalization challenge addressed in this study.
Among the GP-based variants, the RL-based strategy for determining the number of retained cuts (\emph{GPRL}) consistently underperforms the other approaches, showing that using RL solely for cut quantity decisions remains ineffective, even when combined with a learned GP ranking function. Selecting only non-parallel cuts (\emph{GPNP}), which is close to the strategy employed by SCIP, is generally outperformed by \emph{GPSCS}, although the difference remains limited, with \emph{GPNP} requiring only $5\%$ more solving time on average. As reported in Appendix~E, the statistics-based cutoff mechanism reduces the number of selected cuts by approximately $10\%$ compared to retaining all non-parallel cuts, suggesting that a fraction of these additional non-parallel cuts provides limited benefit for the solving process. Applying the statistics-based cutoff to the complete ranked cut list (\emph{GPSAll}) also achieves competitive results but remains slightly below \emph{GPSCS} and \emph{GPNP}. Overall, \emph{GPSCS} provides the best performance among the GP-based variants.
Compared with the other approaches, \emph{GPSCS} also achieves the best overall performance, being the fastest method on three out of seven problem classes and obtaining the lowest average solving time across all classes. It improves over \emph{SCIP} by $6\%$ on average, despite \emph{SCIP} being a highly optimized solver that jointly exploits many handcrafted components, while the GP functions are trained on instances that are on average 41 times smaller than those used for evaluation. 
The transferability experiment further confirms the importance of learning class-specific scoring functions. The \emph{WGPSCS} variant, which randomly applies a GP function trained on a different problem class, consistently achieves the worst performance, showing that the learned scoring functions are effective precisely because they capture characteristics specific to their target problem classes.
Finally, \emph{HEM} also exhibits limited performance. Except on MAXSAT, its solving times are generally at least twice as large as those of the best-performing methods and are often comparable to the \emph{NoCuts} baseline (with the exception of GISP). Moreover, whenever \emph{NoCuts} reaches the time limit of 500 seconds, \emph{HEM} also fails to solve the instances, even in cases where \emph{GPSCS} and \emph{SCIP} solve them within 25 seconds. This suggests potential convergence limitations of the RL-based framework in these settings.

\section{Conclusions}\label{Conclusion}

In this work, we investigated the problem of cut selection, a key component governing the efficiency of MILP solvers. We introduced GPSCS, a fully explicit and human-readable cut strategy that jointly addresses cut ranking and cut quantity decisions. Cut relevance is modeled through nonlinear scoring functions combining cut-specific and contextual features, while selection is further refined by prioritizing non-parallel cuts and dynamically determining the number of cuts to add via a statistical dropout detection mechanism applied to score distributions.

Extensive experimental results demonstrate that GPSCS outperforms the default SCIP configuration and the most practically applicable NN-based baseline for cut selection. Notably, our experiments also reveal potential convergence issues in the NN based RL framework, highlighting limitations induced by delayed and sparse reward signals in large decision spaces.

From the perspective of practical deployment, our contributions can be integrated at two distinct levels. First, the proposed statistical rule for dynamically determining the number of cuts is solver-agnostic and independent of the problem class; it can therefore be directly incorporated into existing MILP solvers, with possible refinements. Second, the GP-based scoring mechanism learns instance-class-specific cut ranking functions, making it particularly suitable as a tuned cutting strategy for researchers or practitioners who repeatedly solve large collections of instances from the same problem family.

Overall, this paper demonstrates that embedding decades of expert knowledge on cutting planes into the design space of learning algorithms enables the discovery of high-performing and interpretable strategies with faster and more stable convergence. Our results highlight population-based methods, such as GP, as a compelling and complementary alternative to prevailing NN-based approaches for improving solver performance.

\appendices


The appendices provide additional material supporting the experimental study of Section~5. In particular, Appendix~\ref{manual_number_of_cuts} reports a separate complementary experiment, in which we extend the GP-based baselines by considering a variant with a fixed number of cuts per separation round. The remaining appendices do not introduce new experiments, but rather offer further insights and analyses derived from the experiments of Section~5. Appendix~\ref{illustrative_example} provides illustrative visualizations of cut score distributions and the resulting decisions on the number of cuts to retain. Appendix~\ref{comvergence_plot_sec} presents the convergence plots of the retained GP scoring functions for the GPSCS method across all problem classes.  Appendix~\ref{contrib_features} analyzes the frequency of feature usage in the generated GP scoring functions.  Finally, Appendix~\ref{nb_of_cuts} summarizes, for each problem class, the average number of cuts selected by all baselines including GP.

\section{Complementary Experiments: Fixed Number of Cuts for GP}\label{manual_number_of_cuts}

We evaluate GP-based scoring functions trained with a fixed number of cuts per separation round, combined with the parallelism-filtering procedure. 

We focus on the three synthetic instance classes \textsc{FCMCNF}, \textsc{GISP}, and \textsc{MAXSAT}. For each class, GP is trained using the same hyperparameters as in the main experiments, while varying the fixed cut number retained, from $5$ to $70$. For each setting, 15 independent GP runs with different random seeds are performed, yielding 15 distinct scoring functions per cut number and class.

\begin{figure}
    \centering
    \includegraphics[width=1\linewidth]{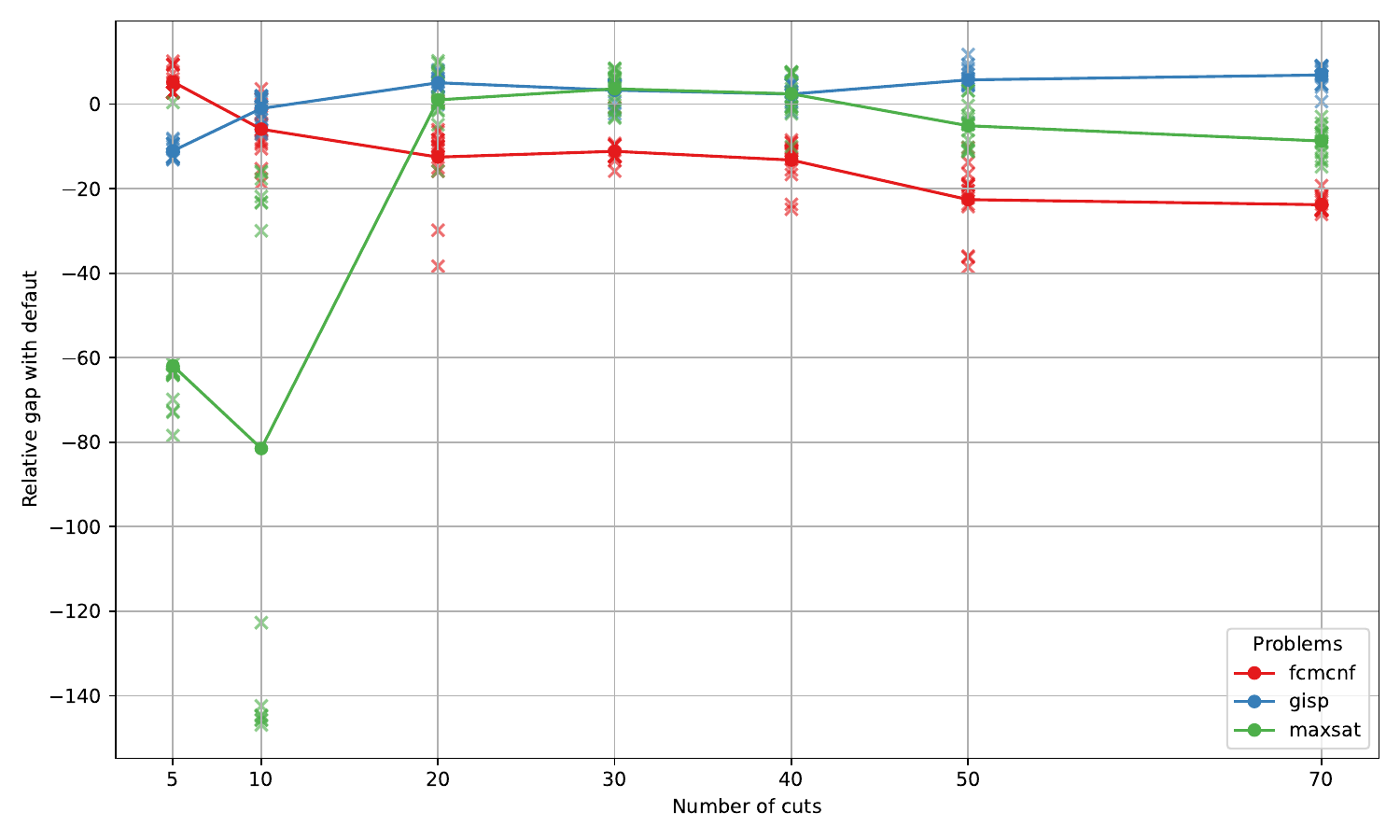}
    \caption{Relative improvement in solving time over the SCIP baseline (y-axis) for GP-based scoring functions combined with a fixed number of cuts per separation round (x-axis).}
    \label{fix_nb_of_cuts}
\end{figure}

Each scoring function is evaluated on its corresponding test set. \cref{fix_nb_of_cuts} reports the relative improvement in solving time over the default SCIP configuration, with crosses indicating individual runs and the line representing the mean performance across the 15 seeds.

Two main observations emerge. First, the optimal fixed number of cuts differs substantially across classes: around 5 for \textsc{FCMCNF}, 20–30 for \textsc{MAXSAT}, and 50–70 for \textsc{GISP}, confirming that no single fixed cut budget is universally effective.

Second, all fixed-cut variants are outperformed by the adaptive dropout-based cut quantity mechanism used in GPSCS, which yields relative improvements of 8.4\%, 12\%, and 2.5\% for \textsc{FCMCNF}, \textsc{GISP}, and \textsc{MAXSAT}, respectively. This highlights the importance of dynamically adjusting the number of cuts both across instances and during the branch-and-bound process.

Overall, these results further support the need for an adaptive cut quantity strategy and validate the benefit of jointly learning cut ranking while dynamically determining how many cuts to apply at each separation round.

\section{Cut Scoring Representation and Selection of the Final Number of Cuts}~\label{illustrative_example}

\begin{figure*}[t]
    \centering
    \begin{minipage}{0.345\textwidth}
        \centering
        \includegraphics[width=\linewidth]{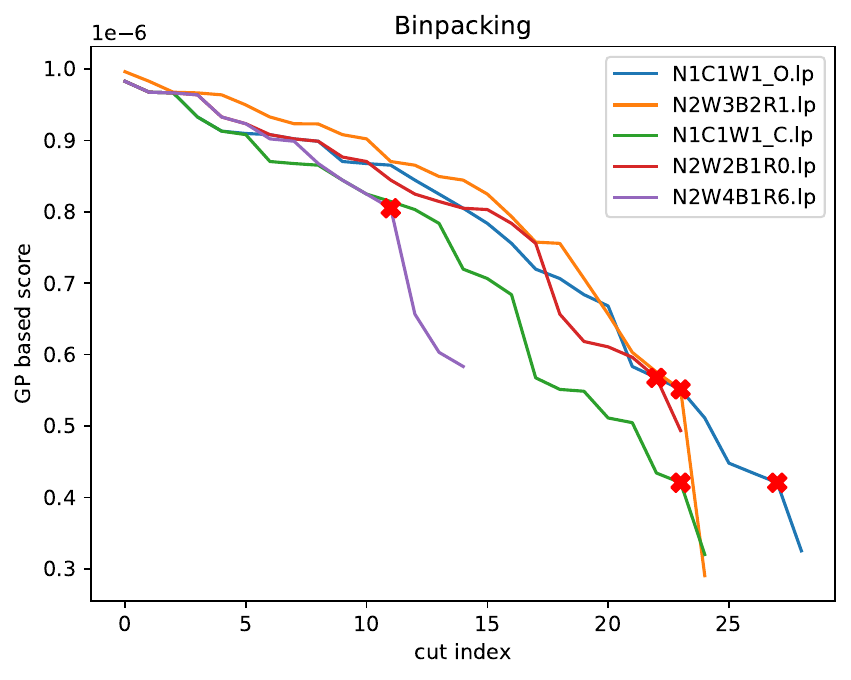}
    \end{minipage}
    \begin{minipage}{0.32\textwidth}
        \centering
        \includegraphics[width=\linewidth]{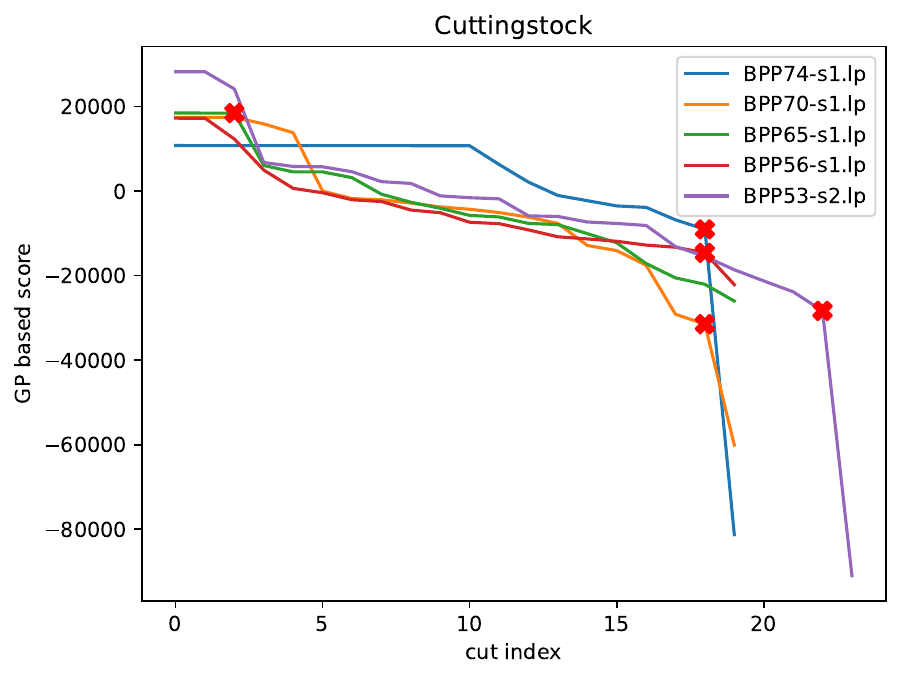}
    \end{minipage}
    \begin{minipage}{0.32\textwidth}
        \centering
        \includegraphics[width=\linewidth]{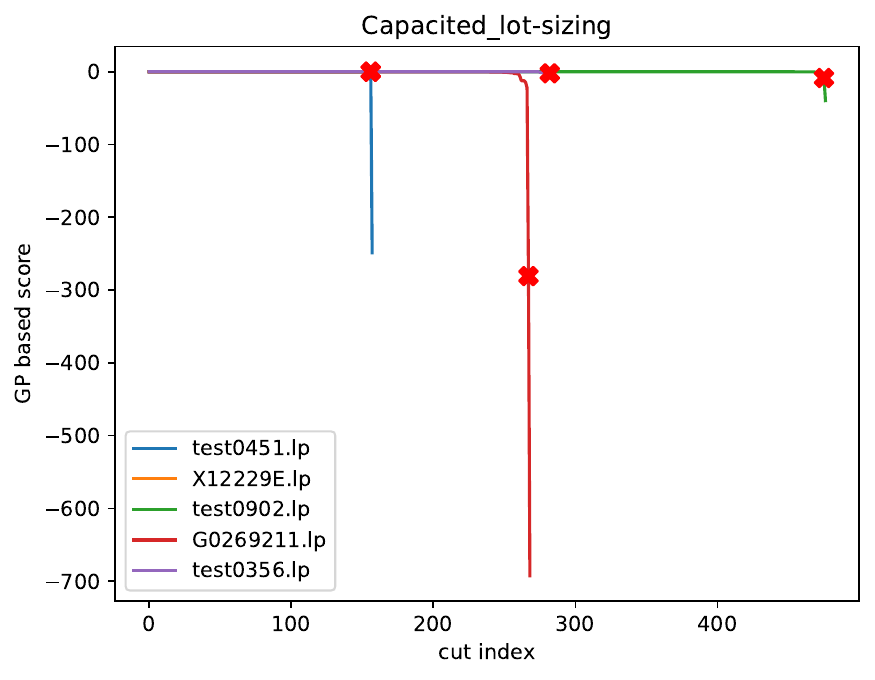}
    \end{minipage}
    \caption{Scores for cuts added during a cut generation round at the root node for 5 instances in each StrIPlib problem class. Red crosses indicate where the heuristic stops adding cuts.}
\label{score_distribution}
\end{figure*}

\begin{figure*}[t]
    \centering
    \begin{minipage}{0.345\textwidth}
        \centering
        \includegraphics[width=\linewidth]{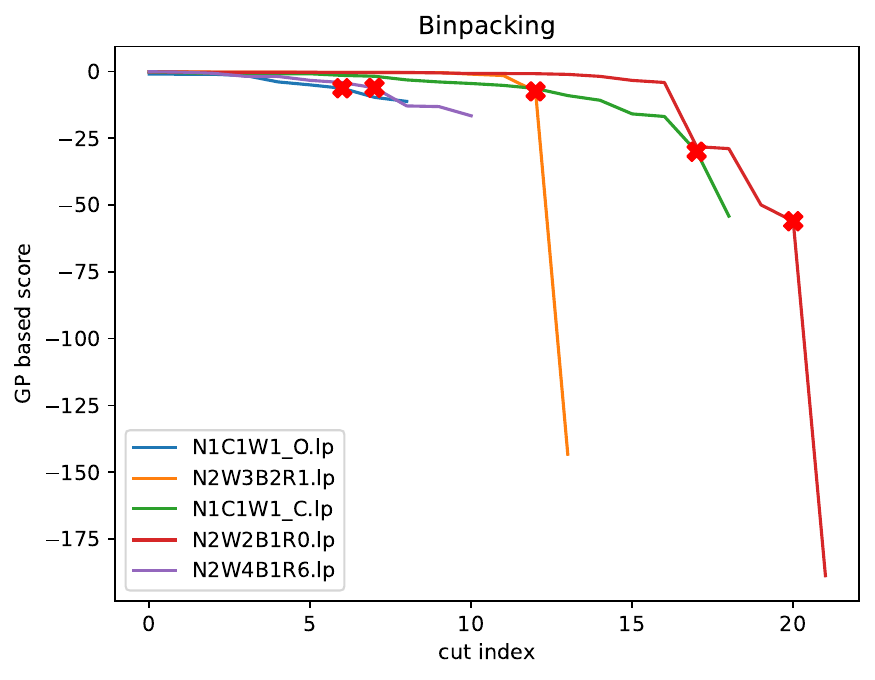}
    \end{minipage}
    \begin{minipage}{0.32\textwidth}
        \centering
        \includegraphics[width=\linewidth]{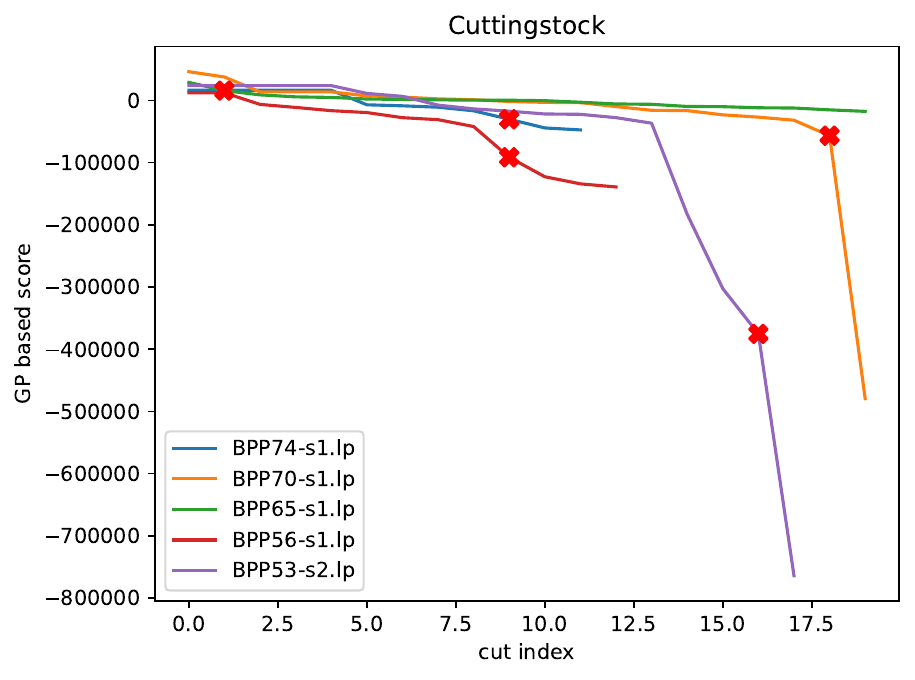}
    \end{minipage}
    \begin{minipage}{0.32\textwidth}
        \centering
        \includegraphics[width=\linewidth]{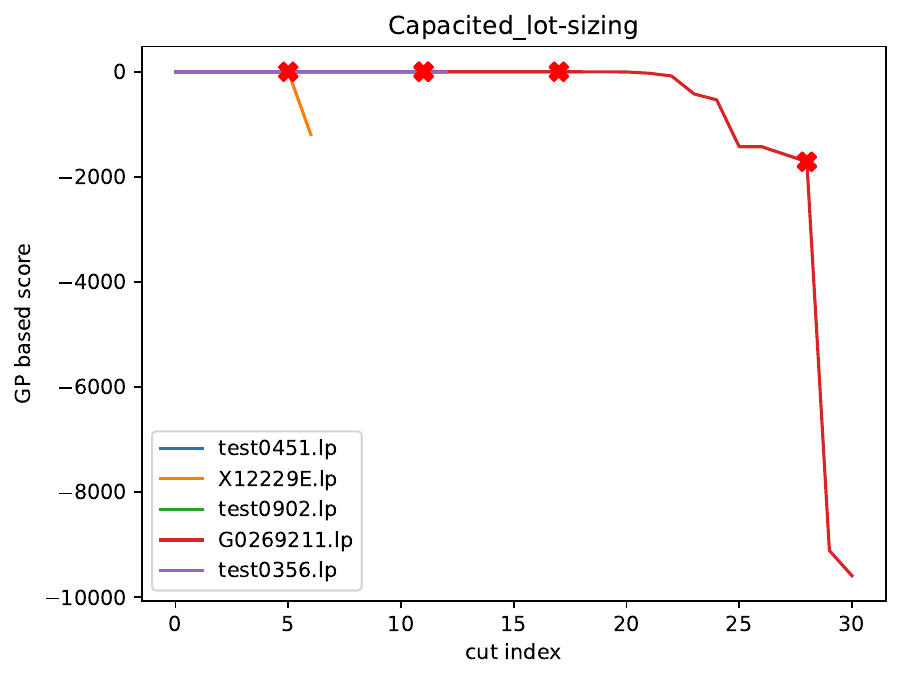}
    \end{minipage}
    \caption{Scores for cuts added at a first non-root node, same setup as in~\cref{score_distribution}.}
    \label{score_distribution_non_root}
\end{figure*}

After applying the scoring function and filtering for parallel cuts, the final number of cuts to add is determined using the cutoff detection procedure. \cref{score_distribution} shows the distribution of filtered scores at the root node for 5 instances from the StrIPlib classes~\cite{bastubbe_striplib_2025} Binpacking, Cuttingstock, and Capacitated Lot-Sizing, along with the number of cuts ultimately added. \cref{score_distribution_non_root} shows the corresponding distributions for a first non-root node.

The score distributions vary substantially across both problem classes and node types (root versus non-root). Root-node distributions generally exhibit lower variability, particularly for Binpacking and Capacitated Lot Sizing, where most instances display a single pronounced score drop near the end of the ranking. The number of generated non-parallel cuts also differs considerably across classes, ranging from more than 100 cuts at the root node for Capacitated Lot Sizing to fewer than 40 in the other settings.
Overall, the proposed dropout detection mechanism accurately identifies the main breakpoints in the score distributions. This behavior is especially evident for Binpacking and Capacitated Lot Sizing, whose score profiles are relatively stable across cuts with one big drop. Although Cuttingstock exhibits greater variability, a pronounced score drop is still observed near the end of the ranking in most instances and is successfully detected by the proposed method.

\section{Convergence Plots of the GP}~\label{comvergence_plot_sec}

\begin{figure*}[t]
    \centering

    \begin{minipage}{0.32\textwidth}
        \centering
        \includegraphics[width=\linewidth]{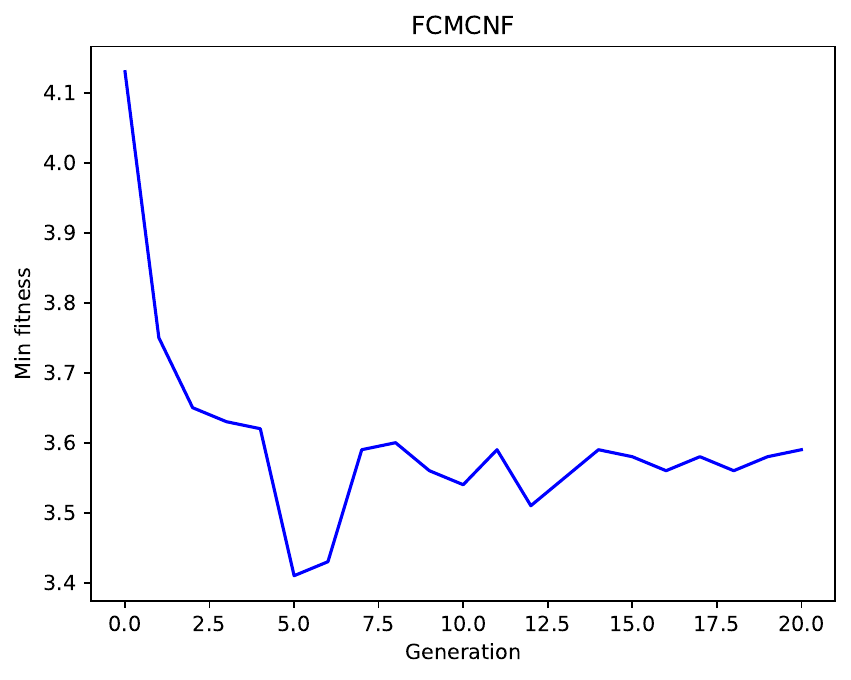}
    \end{minipage}
    \begin{minipage}{0.32\textwidth}
        \centering
        \includegraphics[width=\linewidth]{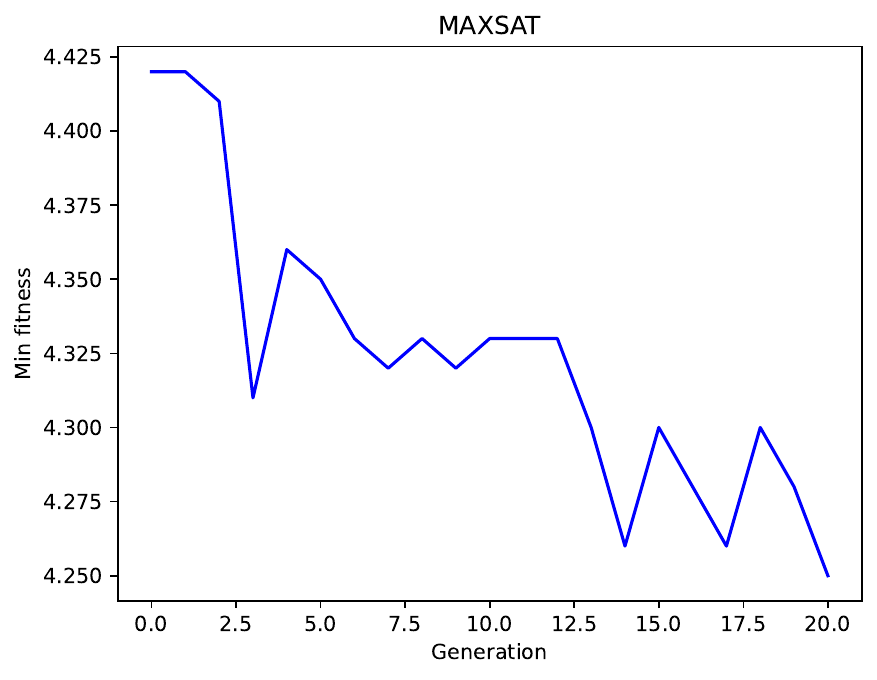}
    \end{minipage}
    \begin{minipage}{0.32\textwidth}
        \centering
        \includegraphics[width=\linewidth]{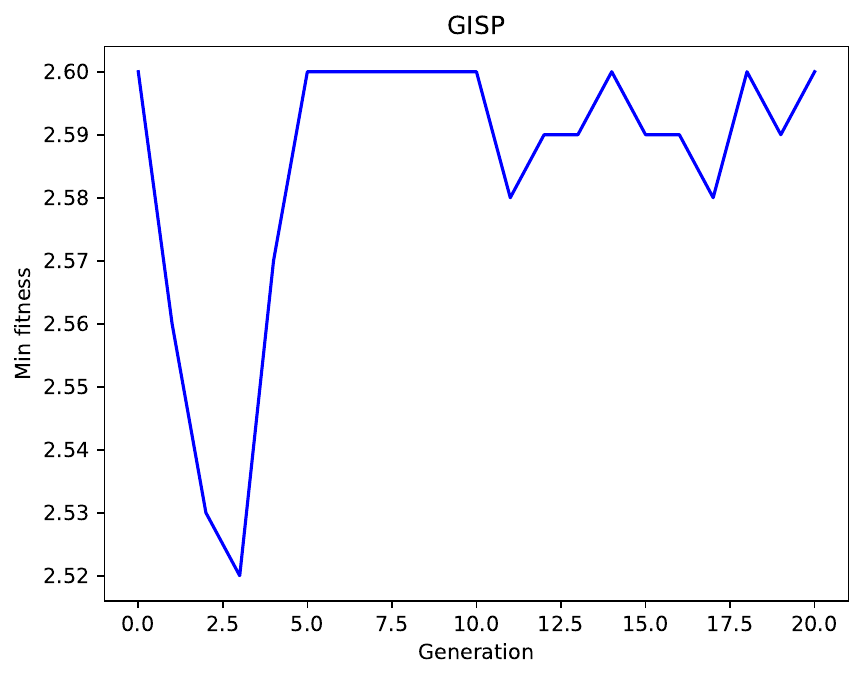}
    \end{minipage}
    \begin{minipage}{0.32\textwidth}
        \centering
        \includegraphics[width=\linewidth]{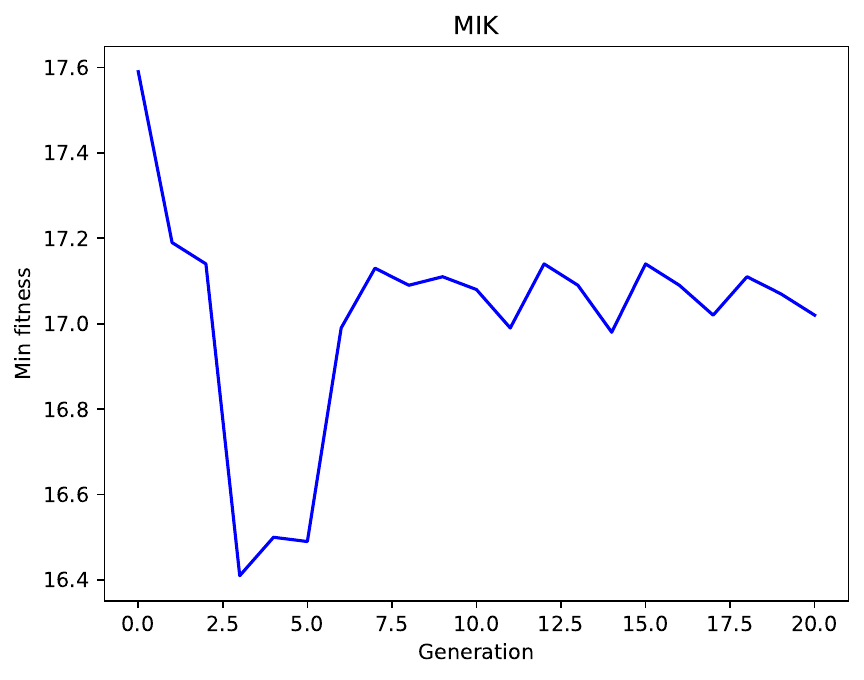}
    \end{minipage}

    \begin{minipage}{0.32\textwidth}
        \centering
        \includegraphics[width=\linewidth]{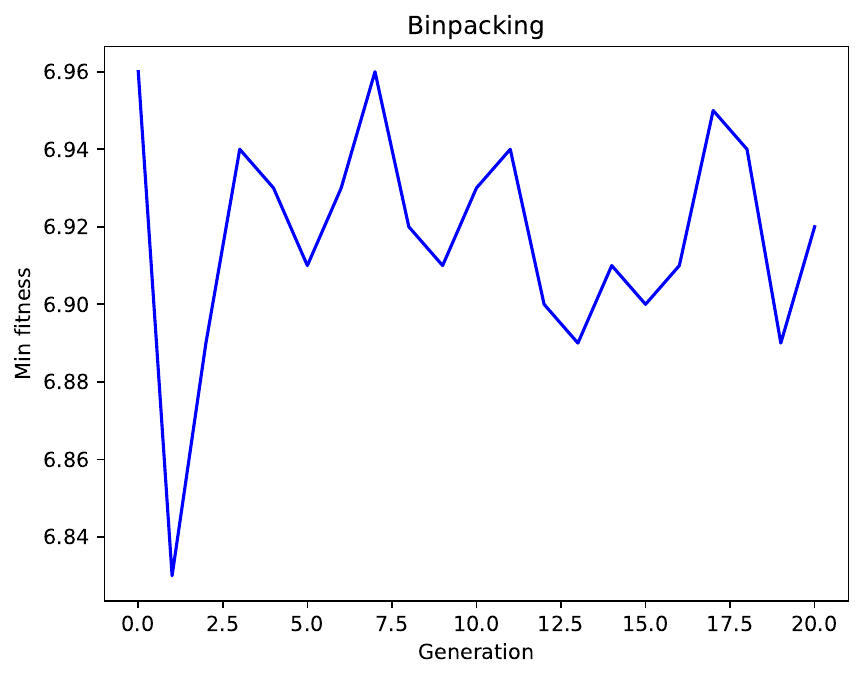}
    \end{minipage}
    \begin{minipage}{0.32\textwidth}
        \centering
        \includegraphics[width=\linewidth]{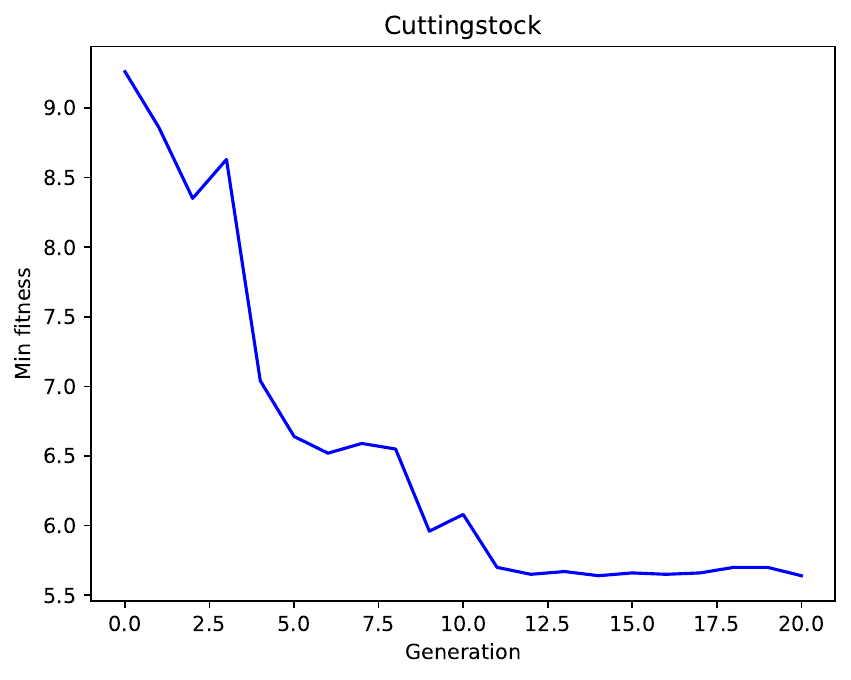}
    \end{minipage}
    \begin{minipage}{0.32\textwidth}
        \centering
        \includegraphics[width=\linewidth]{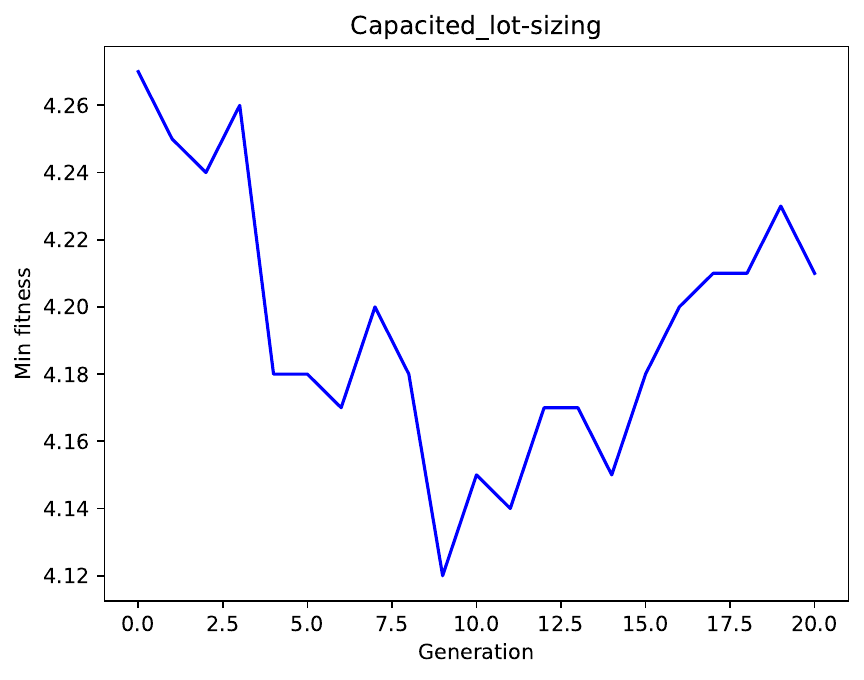}
    \end{minipage}

    \caption{Convergence of GP training: best (minimum) solving time in the population at each generation over 20 generations, evaluated on the training set.}
    \label{convergence_plot}
\end{figure*}

For each problem class, we perform 15 independent GP training runs and retain the model with the best performance on the training set. \cref{convergence_plot} shows the convergence of these selected GPs, reporting at each generation the best individual in the population, i.e., the scoring function achieving the lowest average solving time on the training set.

Convergence patterns differ across classes. For most classes, especially the three synthetic ones, the binpacking and the capacitated lot-sizing instances, the training process is largely monotonic, with improvements of from 0.2 to 0.7 seconds over 20 generations due to the large population size and short solving times. For the remaining classes, the GP evolution achieves more substantial improvements, reducing average solving times on the training set by more than one second over the course of training. Nevertheless, in all cases, the convergence curves remain largely monotonic, clearly illustrating the consistent contribution of GP-driven exploration. Overall, the convergence curves exhibit a clear monotonic trend, demonstrating the effectiveness of GP-driven search. At the same time, the variability across runs indicates that the quality of the initial population strongly influences the final improvement achieved by the evolutionary process.

\section{Feature Contribution of the GP Fonctions}\label{contrib_features}

For each of the five GP-based alternatives (\emph{GPSCS}, \emph{GPCS}, \emph{GPNP}, \emph{GPSNP}, and \emph{GPRL}), 15 random seeds are used to generate 15 GP scoring functions, resulting in a total of 75 functions. \cref{table_features_grouped} reports the five most frequently occurring features among these functions, distinguishing whether each feature contributes positively to the score when its value increases (\emph{Increase}) or negatively when its value increases (\emph{Decrease}).
A first observation is that $eff$ consistently appears among the most influential features across almost all problem classes and GP variants, particularly in the \emph{Increase} category (with the exception of \emph{GISP}). This suggests that cut efficacy remains a key and robust indicator of cut quality, independently of the problem class and the selection strategy. Beyond this common trend, the remaining top features exhibit substantial variability across classes, both in terms of feature identity and their contribution direction. For instance, FCMCNF shows the strongest consensus, with the same feature appearing 60 times in the \emph{Increase} category, whereas the corresponding consensus is much weaker for MIK, with only 26 occurrences.
These results also highlight that some features appear to be particularly relevant for specific problem classes. For example, $N_{\text{nonz}}$ appears 30 times in the \emph{Increase} category for one class, while occurring fewer than five times in the GP functions learned for all other classes. This suggests that, although some cut characteristics such as efficacy generalize broadly, other features capture problem-specific properties that are exploited by the learned scoring functions.

\begin{table}[htbp]
\centering
\footnotesize
\begin{tabular}{l|c|c}
\textbf{Feature} & \textbf{Increase} & \textbf{Decrease} \\
\hline
\multicolumn{3}{l}{\textbf{binpacking}} \\
\hline
$eff$ & 34 & 10 \\
$d_i$ & 19 & 14 \\
$dcd$ & 20 & 1 \\
$M$ & 5 & 10 \\
$nvi$ & 7 & 6 \\
\hline
\multicolumn{3}{l}{\textbf{capacited\_lot-sizing}} \\
\hline
$eff$ & 44 & 1 \\
$dcd$ & 32 & 0 \\
$nvi$ & 0 & 12 \\
$N_{\text{nonz}}$ & 2 & 3 \\
$obp$ & 1 & 3 \\
\hline
\multicolumn{3}{l}{\textbf{cuttingstock}} \\
\hline
$obp$ & 17 & 28 \\
$dcd$ & 33 & 4 \\
$M$ & 12 & 5 \\
$eff$ & 11 & 4 \\
$d_i$ & 10 & 4 \\
\hline
\multicolumn{3}{l}{\textbf{fcmcnf}} \\
\hline
$eff$ & 59 & 6 \\
$nvi$ & 5 & 17 \\
$M$ & 8 & 5 \\
$n$ & 8 & 4 \\
$m$ & 4 & 4 \\
\hline
\multicolumn{3}{l}{\textbf{GISP}} \\
\hline
$N_{\text{nonz}}$ & 30 & 0 \\
$obp$ & 23 & 3 \\
$nvi$ & 20 & 3 \\
$M$ & 6 & 9 \\
$d_i$ & 12 & 2 \\
\hline
\multicolumn{3}{l}{\textbf{MIK}} \\
\hline
$dcd$ & 26 & 0 \\
$M$ & 22 & 0 \\
$eff$ & 20 & 1 \\
$obp$ & 15 & 0 \\
$m$ & 7 & 0 \\
\hline
\multicolumn{3}{l}{\textbf{WPSM}} \\
\hline
$nvi$ & 43 & 5 \\
$eff$ & 29 & 17 \\
$obp$ & 11 & 22 \\
$dcd$ & 13 & 1 \\
$m$ & 8 & 3 \\
\end{tabular}
\caption{Feature contributions (increase/decrease) of the GP fonctions per problem class.}
\label{table_features_grouped}
\end{table}

\section{Mean Number of Cuts by Strategy}~\label{nb_of_cuts}

\begin{table*}[h]
\centering
\footnotesize
\begin{tabular}{lccccccc|c}
\hline
Method & FCMCNF & MAXSAT & GISP & MIK & Binpacking & Cuttingstock & Cap lot-sizing & Mean\\
\hline
GPRL & $44.5 \pm 15.0$ & $46.7 \pm 16.9$ & $4.5 \pm 0.4$ & $21.1 \pm 10.5$ & $67.8 \pm 51.4$ & $118.2 \pm 116.6$ & $17.7 \pm 8.4$ & 45.79 \\
GPAll & $45.3 \pm 12.5$ & $66.7 \pm 16.5$ & $24.0 \pm 1.7$ & $56.2 \pm 27.6$ & $124.1 \pm 137.3$ & $124.2 \pm 134.8$ & $20.3 \pm 9.5$ & 65.83 \\
GPNP & $13.7 \pm 4.3$ & $25.4 \pm 11.1$ & $2.4 \pm 0.2$ & $7.4 \pm 2.4$ & $12.0 \pm 8.5$ & $14.8 \pm 8.6$ & $4.0 \pm 1.6$ & 11.39 \\
GPSAll & $21.6 \pm 9.6$ & $37.9 \pm 18.2$ & $23.1 \pm 2.0$ & $11.8 \pm 3.8$ & $15.7 \pm 8.1$ & $71.4 \pm 90.3$ & $5.1 \pm 2.1$ & 26.66 \\
GPSCS & $12.6 \pm 3.5$ & $20.6 \pm 8.8$ & $1.6 \pm 0.1$ & $8.0 \pm 3.6$ & $9.6 \pm 7.4$ & $13.6 \pm 8.5$ & $3.6 \pm 1.5$ & 9.94 \\
\hline
\end{tabular}
\caption{Mean number of cuts per separation round across instances for each method and problem class.}
\label{perf_table_appendix}
\end{table*}

For the five GP-based alternatives, the main difference lies in the number of cuts selected during GP training, which in turn influences the final scoring function produced by the GP. \cref{perf_table_appendix} reports the average number of cuts added per instance class for each method. Overall, \emph{GPSCS} consistently selects fewer cuts than \emph{GPNP}, removing approximately $10\%$ of the non-parallel cuts on average. This behavior can also be observed qualitatively in the score distributions shown in \cref{score_distribution,score_distribution_non_root}. The only exception is MIK, where \emph{GPSCS} retains slightly more cuts than \emph{GPNP}. This is likely explained by differences in the GP-learned ranking functions, which induce different B\&B search trees and consequently generate different sets of candidate cuts. In contrast, the difference in the number of retained cuts between \emph{GPSCS} and \emph{GPAll} is substantially larger, mirroring the larger performance gap observed between these two methods in terms of solving time.



\bibliographystyle{ieeetr}
\bibliography{references}

@misc{gurobi_gurobi_2024,
	title = {Gurobi {Optimizer} {Reference} {Manual}},
	url = {https://docs.gurobi.com/projects/optimizer/en/current/index.html},
	urldate = {2026-01-15},
	author = {gurobi},
	year = {2024},
}

@misc{ibm_cplex_ibm_2022,
	title = {{IBM} {ILOG} {CPLEX} {Optimization} {Studio}},
	copyright = {© Copyright IBM Corporation 2022},
	url = {https://www.ibm.com/docs/en/icos/22.1.1?topic=optimizers-users-manual-cplex},
	language = {en-US},
	urldate = {2026-01-15},
	author = {IBM CPLEX},
	month = dec,
	year = {2022},
}

@inproceedings{wang_learning_2023,
	title = {Learning {Cut} {Selection} for {Mixed}-{Integer} {Linear} {Programming} via {Hierarchical} {Sequence} {Model}},
	url = {https://openreview.net/forum?id=Zob4P9bRNcK},
	language = {en},
	urldate = {2025-06-20},
	booktitle = {The {Eleventh} {International} {Conference} on {Learning} {Representations}},
	publisher = {The Eleventh International Conference on Learning Representations},
	author = {Wang, Zhihai and Li, Xijun and Wang, Jie and Kuang, Yufei and Yuan, Mingxuan and Zeng, Jia and Zhang, Yongdong and Wu, Feng},
	month = sep,
	year = {2023},
}

@article{atamturk_facets_2003,
	title = {On the facets of the mixed–integer knapsack polyhedron},
	volume = {98},
	issn = {1436-4646},
	url = {https://doi.org/10.1007/s10107-003-0400-z},
	doi = {10.1007/s10107-003-0400-z},
	language = {en},
	number = {1},
	urldate = {2026-01-14},
	journal = {Mathematical Programming},
	author = {Atamturk, Alper},
	month = sep,
	year = {2003},
	pages = {145--175},
}

@book{norman_pdq_2003,
	title = {{PDQ} {Statistics}},
	isbn = {978-1-60795-077-6},
	language = {en},
	publisher = {PMPH USA},
	author = {Norman, Geoffrey R. and PhD, David L. Streiner},
	month = jan,
	year = {2003},
	note = {Google-Books-ID: Lw\_2DwAAQBAJ},
}

@inproceedings{tang_reinforcement_2020,
	title = {Reinforcement {Learning} for {Integer} {Programming}: {Learning} to {Cut}},
	issn = {2640-3498},
	shorttitle = {Reinforcement {Learning} for {Integer} {Programming}},
	url = {https://proceedings.mlr.press/v119/tang20a.html},
	language = {en},
	urldate = {2026-01-07},
	booktitle = {Proceedings of the 37th {International} {Conference} on {Machine} {Learning}},
	publisher = {PMLR},
	author = {Tang, Yunhao and Agrawal, Shipra and Faenza, Yuri},
	month = nov,
	year = {2020},
	pages = {9367--9376},
}

@inproceedings{paulus_learning_2022,
	title = {Learning to {Cut} by {Looking} {Ahead}: {Cutting} {Plane} {Selection} via {Imitation} {Learning}},
	issn = {2640-3498},
	shorttitle = {Learning to {Cut} by {Looking} {Ahead}},
	url = {https://proceedings.mlr.press/v162/paulus22a.html},
	language = {en},
	urldate = {2026-01-07},
	booktitle = {Proceedings of the 39th {International} {Conference} on {Machine} {Learning}},
	publisher = {PMLR},
	author = {Paulus, Max B. and Zarpellon, Giulia and Krause, Andreas and Charlin, Laurent and Maddison, Chris},
	month = jun,
	year = {2022},
	pages = {17584--17600},
}

@misc{huang_learning_2021,
	title = {Learning to {Select} {Cuts} for {Efficient} {Mixed}-{Integer} {Programming}},
	url = {http://arxiv.org/abs/2105.13645},
	doi = {10.48550/arXiv.2105.13645},
	urldate = {2026-01-07},
	publisher = {arXiv},
	author = {Huang, Zeren and Wang, Kerong and Liu, Furui and Zhen, Hui-ling and Zhang, Weinan and Yuan, Mingxuan and Hao, Jianye and Yu, Yong and Wang, Jun},
	month = oct,
	year = {2021},
	note = {arXiv:2105.13645 [math]},
}

@article{gilmore_linear_1961,
	title = {A {Linear} {Programming} {Approach} to the {Cutting}-{Stock} {Problem}},
	copyright = {© 1961 INFORMS},
	url = {https://pubsonline.informs.org/doi/abs/10.1287/opre.9.6.849},
	doi = {10.1287/opre.9.6.849},
	language = {en},
	urldate = {2026-01-06},
	journal = {Operations Research},
	publisher = {INFORMS},
	author = {Gilmore, P. C. and Gomory, R. E.},
	month = dec,
	year = {1961},
}

@misc{suhl_implementing_2012,
	title = {Implementing cutting plane management and selection techniques},
	url = {https://optimization-online.org/?p=12261},
	language = {en-US},
	urldate = {2026-01-05},
	publisher = {Optimization Online},
	author = {Suhl, U. H. and Wesselmann, Franz},
	month = dec,
	year = {2012},
}

@article{rais_operations_2011,
	title = {Operations {Research} in {Healthcare}: a survey},
	volume = {18},
	copyright = {© 2010 The Authors. International Transactions in Operational Research © 2010 International Federation of Operational Research Societies},
	issn = {1475-3995},
	shorttitle = {Operations {Research} in {Healthcare}},
	url = {https://onlinelibrary.wiley.com/doi/abs/10.1111/j.1475-3995.2010.00767.x},
	doi = {10.1111/j.1475-3995.2010.00767.x},
	language = {en},
	number = {1},
	urldate = {2024-03-04},
	journal = {International Transactions in Operational Research},
	author = {Rais, Abdur and Viana, Ana},
	year = {2011},
	pages = {1--31},
}

@article{maudet_search_2025,
	title = {Search {Strategy} {Generation} for {Branch} and {Bound} {Using} {Genetic} {Programming}},
	volume = {39},
	copyright = {Copyright (c) 2025 Association for the Advancement of Artificial Intelligence},
	issn = {2374-3468},
	url = {https://ojs.aaai.org/index.php/AAAI/article/view/33229},
	doi = {10.1609/aaai.v39i11.33229},
	language = {en},
	number = {11},
	urldate = {2025-08-13},
	journal = {Proceedings of the AAAI Conference on Artificial Intelligence},
	author = {Maudet, Gwen and Danoy, Grégoire},
	month = apr,
	year = {2025},
	pages = {11299--11308},
}

@article{dey_theoretical_2018,
	title = {Theoretical challenges towards cutting-plane selection},
	volume = {170},
	issn = {1436-4646},
	url = {https://doi.org/10.1007/s10107-018-1302-4},
	doi = {10.1007/s10107-018-1302-4},
	language = {en},
	number = {1},
	urldate = {2025-06-20},
	journal = {Mathematical Programming},
	author = {Dey, Santanu S. and Molinaro, Marco},
	month = jul,
	year = {2018},
	pages = {237--266},
}

@inproceedings{cappart_combinatorial_2021,
	address = {Montreal, Canada},
	title = {Combinatorial {Optimization} and {Reasoning} with {Graph} {Neural} {Networks}},
	isbn = {978-0-9992411-9-6},
	url = {https://www.ijcai.org/proceedings/2021/595},
	doi = {10.24963/ijcai.2021/595},
	language = {en},
	urldate = {2025-06-20},
	booktitle = {Proceedings of the {Thirtieth} {International} {Joint} {Conference} on {Artificial} {Intelligence}},
	publisher = {International Joint Conferences on Artificial Intelligence Organization},
	author = {Cappart, Quentin and Chételat, Didier and Khalil, Elias B. and Lodi, Andrea and Morris, Christopher and Veličković, Petar},
	month = aug,
	year = {2021},
	pages = {4348--4355},
}

@article{scavuzzo_machine_2024,
	title = {Machine learning augmented branch and bound for mixed integer linear programming},
	issn = {0025-5610, 1436-4646},
	url = {https://link.springer.com/10.1007/s10107-024-02130-y},
	doi = {10.1007/s10107-024-02130-y},
	language = {en},
	urldate = {2025-06-11},
	journal = {Mathematical Programming},
	author = {Scavuzzo, Lara and Aardal, Karen and Lodi, Andrea and Yorke-Smith, Neil},
	month = aug,
	year = {2024},
}

@misc{bastubbe_striplib_2025,
	title = {{strIPlib}: {Structured} {Integer} {Programming} {Library}},
	url = {https://striplib.or.rwth-aachen.de},
	urldate = {2024-10-21},
	author = {Bastubbe, Michael and Helber, Alexander and Kirchhart, Lukas and L\{{\textbackslash}"u\}bbecke, Marco and Rieken, Niklas and Witt, Jonas},
	year = {2025},
}

@article{turner_adaptive_2023,
	title = {Adaptive {Cut} {Selection} in {Mixed}-{Integer} {Linear} {Programming}},
	volume = {4},
	issn = {2777-5860},
	url = {https://ojmo.centre-mersenne.org/articles/10.5802/ojmo.25/},
	doi = {10.5802/ojmo.25},
	urldate = {2025-01-22},
	journal = {Open Journal of Mathematical Optimization},
	author = {Turner, Mark and Koch, Thorsten and Serrano, Felipe and Winkler, Michael},
	year = {2023},
	pages = {1--28},
}

@article{bengio_machine_2021,
	title = {Machine {Learning} for {Combinatorial} {Optimization}: a {Methodological} {Tour} d'{Horizon}},
	shorttitle = {Machine {Learning} for {Combinatorial} {Optimization}},
	url = {http://arxiv.org/abs/1811.06128},
	language = {en},
	urldate = {2024-01-30},
	journal = {European Journal of Operational Research},
	author = {Bengio, Yoshua and Lodi, Andrea and Prouvost, Antoine},
	year = {2021},
	note = {arXiv:1811.06128 [cs, stat]},
}

@phdthesis{achterberg_constraint_2007,
	title = {Constraint {Integer} {Programming}},
	language = {de},
	school = {ZIB Berlin},
	author = {Achterberg, Tobias},
	year = {2007},
}

@misc{bolusani_scip_2024,
	title = {The {SCIP} {Optimization} {Suite} 9.0},
	url = {https://optimization-online.org/?p=25734},
	language = {en-US},
	urldate = {2024-06-13},
	publisher = {Optimization Online},
	author = {Bolusani, Suresh and Besançon, Mathieu and Bestuzheva, Ksenia and Chmiela, Antonia and Dionisio, Joao and Donkiewicz, Tim and Doornmalen, Jasper van and Eifler, Leon and Ghannam, Mohammed and Gleixner, Ambros and Graczyk, Christoph and Halbig, Katrin and Hedtke, Ivo and Hoen, Alexander and Hojny, Christopher and Hulst, Rolf van der and Kamp, Dominik and Koch, Thorsten and Kofler, Kevin and Lentz, Jurgen and Manns, Julian and Mexi, Gioni and Mühmer, Erik and Pfetsch, Marc E. and Schlösser, Franziska and Serrano, Felipe and Shinano, Yuji and Turner, Mark and Vigerske, Stefan and Weninger, Dieter and Xu, Liding},
	month = feb,
	year = {2024},
}

@article{erdos_evolution_1961,
	title = {On the evolution of random graphs},
	volume = {38},
	url = {https://cir.nii.ac.jp/crid/1572261550756525440},
	urldate = {2024-06-05},
	journal = {Bulletin of the Institute of International Statistics},
	author = {Erdos, P},
	year = {1961},
	pages = {343--347},
}

@inproceedings{bejar_generating_2009,
	title = {Generating {Hard} {Instances} for {MaxSAT}},
	issn = {2378-2226},
	url = {https://ieeexplore.ieee.org/document/5010398},
	doi = {10.1109/ISMVL.2009.58},
	urldate = {2024-05-17},
	booktitle = {2009 39th {International} {Symposium} on {Multiple}-{Valued} {Logic}},
	author = {Béjar, Ramón and Cabiscol, Alba and Manyà, Felip and Planes, Jordi},
	month = may,
	year = {2009},
	pages = {191--195},
}

@misc{chmiela_learning_2021,
	title = {Learning to {Schedule} {Heuristics} in {Branch}-and-{Bound}},
	url = {http://arxiv.org/abs/2103.10294},
	doi = {10.48550/arXiv.2103.10294},
	urldate = {2024-05-17},
	publisher = {arXiv},
	author = {Chmiela, Antonia and Khalil, Elias B. and Gleixner, Ambros and Lodi, Andrea and Pokutta, Sebastian},
	month = mar,
	year = {2021},
	note = {arXiv:2103.10294 [cs, math]},
}

@article{hewitt_combining_2009,
	title = {Combining {Exact} and {Heuristic} {Approaches} for the {Capacitated} {Fixed}-{Charge} {Network} {Flow} {Problem}},
	copyright = {Copyright © 2010, INFORMS},
	url = {https://pubsonline.informs.org/doi/abs/10.1287/ijoc.1090.0348},
	doi = {10.1287/ijoc.1090.0348},
	language = {en},
	urldate = {2024-05-17},
	journal = {INFORMS Journal on Computing},
	publisher = {INFORMS},
	author = {Hewitt, Mike and Nemhauser, George L. and Savelsbergh, Martin W. P.},
	month = sep,
	year = {2009},
}

@inproceedings{luke_fighting_2002,
	address = {Berlin, Heidelberg},
	title = {Fighting {Bloat} with {Nonparametric} {Parsimony} {Pressure}},
	isbn = {978-3-540-45712-1},
	doi = {10.1007/3-540-45712-7_40},
	language = {en},
	booktitle = {Parallel {Problem} {Solving} from {Nature} — {PPSN} {VII}},
	publisher = {Springer},
	author = {Luke, Sean and Panait, Liviu},
	editor = {Guervós, Juan Julián Merelo and Adamidis, Panagiotis and Beyer, Hans-Georg and Schwefel, Hans-Paul and Fernández-Villacañas, José-Luis},
	year = {2002},
	pages = {411--421},
}

@article{archetti_optimization_2022,
	title = {Optimization in multimodal freight transportation problems: {A} {Survey}},
	volume = {299},
	issn = {0377-2217},
	shorttitle = {Optimization in multimodal freight transportation problems},
	url = {https://www.sciencedirect.com/science/article/pii/S0377221721006263},
	doi = {10.1016/j.ejor.2021.07.031},
	number = {1},
	urldate = {2024-03-04},
	journal = {European Journal of Operational Research},
	author = {Archetti, Claudia and Peirano, Lorenzo and Speranza, M. Grazia},
	month = may,
	year = {2022},
	pages = {1--20},
}

@article{lodi_learning_2017,
	title = {On learning and branching: a survey},
	volume = {25},
	issn = {1863-8279},
	shorttitle = {On learning and branching},
	url = {https://doi.org/10.1007/s11750-017-0451-6},
	doi = {10.1007/s11750-017-0451-6},
	language = {en},
	number = {2},
	urldate = {2023-12-06},
	journal = {TOP},
	author = {Lodi, Andrea and Zarpellon, Giulia},
	month = jul,
	year = {2017},
	pages = {207--236},
}

@article{dokeroglu_hyper-heuristics_2024,
	title = {Hyper-heuristics: {A} survey and taxonomy},
	volume = {187},
	issn = {03608352},
	shorttitle = {Hyper-heuristics},
	url = {https://linkinghub.elsevier.com/retrieve/pii/S0360835223008392},
	doi = {10.1016/j.cie.2023.109815},
	language = {en},
	urldate = {2023-12-06},
	journal = {Computers \& Industrial Engineering},
	author = {Dokeroglu, Tansel and Kucukyilmaz, Tayfun and Talbi, El-Ghazali},
	month = jan,
	year = {2024},
	pages = {109815},
}

@book{talbi_metaheuristics_2009,
	title = {Metaheuristics: {From} {Design} to {Implementation}},
	isbn = {978-0-470-49690-9},
	shorttitle = {Metaheuristics},
	language = {en},
	publisher = {John Wiley \& Sons},
	author = {Talbi, El-Ghazali},
	month = may,
	year = {2009},
	note = {Google-Books-ID: SIsa6zi5XV8C},
}


\end{document}